# Explainable Multimodal AI for Adaptive Calibration of Archaeological Sensing Workflows

Nevio Dubbini[1], Daniël P. van Helden[2], Claudia Sciuto[3], Martina Naso[3], Arthur Leck[4], Clement Joubert[5], Heeli C. Schechter[6], Rémy Chapoulie[4], Gabriele Gattiglia[3]

[1] Miningful srls, Pisa, Italy, neviod@miningfulstudio.eu
[2] King's College London, UK
[3] MAPPA Lab, University of Pisa, Italy
[4] Archéosciences Bordeaux UMR6034 CNRS- University Bordeaux Montaigne, France
[5] INRIA, University of Bordeaux, France
[6] Hebrew University of Jerusalem, Israel

**Abstract**. This paper presents a multimodal machine-learning framework for calibration monitoring, quality assessment, and adaptive acquisition support in archaeological digitisation workflows. The proposed approach operates across photogrammetric 3D reconstruction, hyperspectral imaging, X-ray fluorescence spectroscopy, and Raman spectroscopy through a unified pipeline combining deterministic quality indicators, statistical feature representations, machine-learning classification, anomaly detection, and explainable artificial intelligence (XAI). Rather than replacing instrument-level calibration, the framework introduces an additional algorithmic layer that evaluates whether acquisitions are statistically consistent, physically plausible, and suitable for downstream multimodal integration. For each sensing modality, acquisitions are represented through structured feature spaces encoding geometric, spectral, spatial, and statistical properties. These representations are used to identify degradation patterns such as reconstruction artefacts, illumination inconsistencies, spectral distortions, detector instability, baseline fluctuations, and low signal-to-noise conditions. Supervised and unsupervised learning methods are combined with XAI techniques to support both automatic discrimination between acceptable and problematic acquisitions and interpretation of the underlying causes of degradation. The framework additionally supports adaptive feedback and resource-aware acquisition strategies by linking feature-space deviations to acquisition-level corrective actions. Experimental results obtained on multimodal archaeological datasets demonstrate that the proposed methodology captures meaningful acquisition variability and enables robust quality assessment across heterogeneous sensing modalities.

**Keywords**: Multimodal AI, Archaeological Digitisation, Explainable AI.

# 1 Introduction

The large-scale digitisation of archaeological artefacts increasingly relies on the integration of sensing technologies to capture geometric, spectral, elemental, and molecular information. Techniques such as photogrammetric 3D reconstruction, hyperspectral imaging (HSI), X-ray fluorescence spectroscopy (XRF), and Raman spectroscopy enable the generation of enriched digital representations supporting standardized documentation (of history and archaeology) and archaeometry, with an eye to classification. Recent applications of artificial intelligence in archaeology have demonstrated the potential for artefact recognition and decision support, including automatic ceramic classification from images (Anichini et al. 2021). Integrating these techniques into (semi-)automated workflows introduces challenges related to calibration, acquisition reliability, quality assessment, and cross-modal consistency.

These challenges are amplified by the variability of archaeological artefacts. Although sensing modalities need and provide internal calibration procedures, these do not guarantee that acquisitions are analytically reliable. The definition of acquisition "quality" is modality-dependent and often application-specific. In the case of 3D modelling, reconstruction quality depends on the intended use of the model (Montusiewicz et al. 2026). Work in cultural heritage digitisation has focused on comparing technologies and evaluating photogrammetric outputs (Menna et al. 2016; Polo et al. 2022). Automatic identification of reconstruction failures, mesh inconsistencies, or topological degradation remain underexplored, despite progress in mesh repair and reconstruction-error detection methods (Charton, Baek, and Kim 2021; Sfikas, Perakis, and Theoharis 2022). No standard exists for assessing quality and error propagation (di Filippo et al. 2024; Sorgente et al. 2023).

Similar challenges affect spectroscopic modalities. In HSI, XRF, and Raman spectroscopy, calibration generally operates at two levels: radiometric or instrumental calibration, which ensures measurements' physical consistency, and analytical calibration, which relates signals to compositional or molecular information (Frahm 2024).

In HSI, problems of spectral distortions (Hubbard et al. 2004; Linderholm et al. 2019) are often mitigated manually through region-of-interest selection and spectral filtering (Grahn and Geladi 2007). The scale and complexity of archaeological spectral datasets, however, motivate the development of automatic quality assessment procedures (Sciuto et al. 2022).

Although portable XRF instruments include internal calibration routines and external verification procedures (Frahm 2024), large-scale automated acquisition workflows require additional mechanisms capable of identifying degraded or unreliable spectra. Recent developments in machine learning and data-driven spectral analysis have demonstrated AI's potential for spectral classification and preprocessing (Al-Tameeni et al. 2026; Andric et al. 2024; Yan 2025), although its integration into archaeological workflows remains limited.

Although reference materials such as polystyrene and calcite are widely used for Raman spectroscopy calibration and spectral validation (Itoh and Hanari 2021; Lellinger et al. 2025), no unified framework exists for automatic quality assessment. Madden et al. (2018) have shown that noise and baseline instability can substantially affect peak interpretation and quantitative measurements, supporting the need for automated approaches.

Existing research has largely addressed quality assessment, calibration, and machine learning separately and within individual sensing modalities. To the best of our knowledge, no unified framework has been proposed for multimodal archaeological digitisation.

This paper presents a unified machine learning framework for automatic calibration monitoring, quality assessment, and adaptive acquisition support. The framework is developed by the AUTOMATA project, which aims to integrate robotics and sensing technologies for the enriched digitisation of archaeological ceramics and lithics. It incorporates photogrammetric 3D modelling, HSI, XRF, and Raman spectroscopies through a structure based on: (i) reference-based validation of expected system behaviour; (ii) construction of structured feature spaces from deterministic quality indicators and statistical descriptors; (iii) exploratory statistical analysis and anomaly detection; (iv) supervised and unsupervised machine learning classification; (v) explainable artificial intelligence (XAI) techniques for interpretation; and (vi) feedback mechanisms linking feature-space deviations to acquisition-level corrective actions.
A central aspect of the proposed methodology is the interpretation of calibration as a statistical consistency problem rather than solely an instrumental procedure. The framework combines physical quality indicators with machine learning and XAI methods to support automatic discrimination and interpretation of the underlying causes of degradation.

# 2 Calibration pipeline

The proposed calibration pipeline is designed as a modality-independent framework for monitoring acquisition quality, detecting calibration inconsistencies, and supporting adaptive multimodal digitisation workflows. Each acquisition is represented as a feature vector combining deterministic quality indicators with statistical descriptors. Calibration is therefore interpreted as a statistical consistency problem: acquisitions are evaluated against empirical reference distributions representing nominal operating conditions. The pipeline is organised into six stages, described in the following.

1. Reference stage. At the beginning of each acquisition session, modality-specific reference measurements are acquired and analysed. The empirical distributions define tolerance intervals and confidence bounds for acceptable acquisition behaviour.
2. Feature-space stage. Each acquisition is transformed into a structured feature representation combining deterministic modality-specific metrics and statistical descriptors, capturing both local and global properties. Across all modalities, outlier-related descriptors are derived through Mahalanobis-distance analysis and density-based anomaly detection methods.
3. Exploratory stage. The exploratory stage investigates the statistical structure of the feature space before classification, through univariate and multivariate statistics, correlation analysis, dimensionality reduction techniques, and outlier detection.
4. Classification Stage. The classification stage introduces an automated decision layer on top of the feature representation, formulated as a supervised learning task. Current experiments primarily adopt binary labels (“good” vs “bad”), although the framework is designed to support multi-class operational decisions. Different classifiers were evaluated for each modality because the extracted feature spaces have different statistical characteristics. Simpler models proved sufficient for some modalities (e.g., 3D), whereas tree-based ensemble methods better captured the complexity of spectral data. Supervised models are complemented by unsupervised methods (Gaussian Mixture Models, DBSCAN/HDBSCAN clustering, Isolation Forest, and One-Class SVMs) used to identify anomalous or previously unseen degradation patterns.
5. XAI stage. The XAI stage introduces interpretability into the calibration framework, explaining which features contribute to the decision process. Global interpretability is obtained through feature-importance analysis, while local interpretability is achieved using SHAP and LIME explanations.

6. Feedback stage. The final stage translates statistical and machine-learning outputs into operational decisions. Each acquisition is assigned to one of four classes: acceptable; automatically correctable; reacquire; human intervention required. Decision-making is based on three complementary indicators: (i) a statistical deviation score $D(x)$; (ii) an anomaly score $A(x)$; (iii) a prediction uncertainty score $U(x)$. The deviation score is computed using Mahalanobis distance relative to the reference distributions, while the anomaly score is obtained through Isolation Forest models trained primarily on acceptable acquisitions. The uncertainty score is defined as: $U(x) = 1 - \max(p)$ where $p$ denotes the predicted class probabilities. These components are combined into a global severity score: $S(x) = \alpha D(x) + \beta A(x) + \gamma U(x)$, where $\alpha$, $\beta$, and $\gamma$ are estimated by a multinomial logistic regression. The feedback stage establishes a mapping between feature space and acquisition control parameters.

The six stages are shared across all sensing modalities. In the following sections, the discussion therefore focuses primarily on modality-specific quality indicators, feature definitions, classifier selection, and operational interpretations.

# 3. Data Collection

The proposed framework was developed and validated using multimodal datasets acquired within the AUTOMATA project together with pre-existing acquisitions produced under comparable experimental conditions. At the current stage, acquisitions are performed manually, and the datasets intentionally preserve variability arising from operator choices, acquisition geometry, illumination conditions, environmental factors, and sensor configuration, to realistically model acquisition variability.

The datasets include both acceptable (“good”) and degraded (“bad”) acquisitions. The binary labels were assigned by domain experts, based on modality-specific quality criteria reflecting acquisition suitability for analysis. Owing to the coarse binary distinction adopted at this stage ("good"/"bad"), consensus was reached without significant ambiguity. Annotations are primarily binary, although the framework is designed to support progressively refined multi-class labels as additional data become available. The dataset of 3D models includes photogrammetric models and some generated with structured light. The current dataset includes 51 reconstructions, including 13 of ceramics and 31 of lithics. The HSI dataset consists of data cubes acquired using a Specim IQ push-broom hyperspectral camera covering the 400–1000 nm spectral range with 204 bands. Acquisitions were performed under controlled illumination conditions using halogen lamps and standard dark/white reference procedures. The dataset includes 513 artefacts, including 421 ceramics and 92 lithics. The XRF dataset consists of spectra acquired using Evident (Olympus) Vanta C-series portable XRF devices. The dataset contains 941 measurements from 159 artefacts, including 119 ceramics and 40 lithics. The Raman dataset consists of spectra acquired using a portable Metrohm i-Raman Plus 785S spectrometer equipped with a 785 nm excitation laser. The dataset includes 201 measurements from 56 artefacts, including 46 ceramics and 10 lithics.
At the current stage, no specific class-balancing strategies have been adopted, as the available datasets are still evolving and their class distributions are expected to change substantially with the progressive acquisition of data under robotic operating conditions. Addressing class imbalance through techniques such as class weighting or data resampling will be considered in future developments. To prevent data leakage during model evaluation, all measurements acquired from the same artefact were assigned to the same data partition (not contributing to both the training and test sets).

# 4. 3D modelling

Each model is represented as a structured feature vector encoding geometric, topological, and distributional properties of the generated mesh. Mesh density is evaluated through face-area statistics, where both mean face area and its variance are used. Topological consistency is assessed through edge classification, distinguishing manifold, boundary, and non-manifold edges according to the number of incident faces. Additional manifoldness constraints are evaluated through vertex-fan analysis, quantifying whether neighbouring faces form continuous surfaces around each vertex. Geometric distortion is analysed using edge-length and triangle-shape descriptors. The framework computes global and local edge-length ratios together with triangle aspect-ratio distributions. Dihedral-angle statistics are used to detect artificial flat regions and surface discontinuities. These geometric and topological indicators are summarised using distributional and local-variability descriptors, following the general feature-space construction defined in Section 2. Correlation structures are additionally considered to capture relationships among different mesh degradation patterns.
Logistic Regression produced the strongest validation results, with an accuracy of 0.8 and an AUC of 0.83 (Table 1), suggesting that the discriminative structure of the present feature space is relatively low-dimensional and approximately linearly separable. Given the limited number of available 3D reconstructions, these results should be regarded as preliminary and primarily demonstrate the feasibility of the proposed feature representation and calibration strategy.
SHAP analysis indicates that triangle aspect ratios, boundary-edge statistics, and dihedral-angle descriptors contribute most strongly to the classification outcomes. These analyses identify triangle aspect ratios, boundary-edge statistics, and dihedral-angle descriptors as the most influential variables contributing to classification outcomes.
The current formulation defines the decision logic of the feedback stage and establishes the mapping between detected degradation patterns and candidate corrective actions. At the present stage, however, this mapping is intended as a decision-support mechanism and has not yet been validated within a fully closed-loop acquisition system. Nevertheless, this mapping is already informative as these features are physically interpretable and directly related to common photogrammetric failure modes: boundary or non-manifold edges may indicate incomplete coverage or reconstruction instability; abnormal triangle aspect ratios may suggest misalignment, insufficient image overlap, or artificial surface generation; and dihedral-angle or face-area anomalies may point to poor local mesh reconstruction.

**Table 1**. Performance evaluation of the 3D Logistic Regression classification model.

| Metric | Accuracy | Precision | Recall | F1 | AUC |
|---|---|---|---|---|---|
| Value | 0.8 | 0.79 | 0.79 | 0.79 | 0.83 |

# 5. Hyperspectral imaging

For HSI, the general feature-space framework is instantiated using spectral, spatial, and illumination-quality indicators. Signal-to-noise ratio (SNR) is estimated using local spectral and spatial statistics, where the signal is computed from neighbourhood intensity averages and noise is approximated through local variance, high-frequency spectral components, or residuals obtained after smoothing procedures such as Savitzky–Golay filtering. Spectral smoothness and continuity are quantified through first- and second-order spectral derivatives, total variation, and local polynomial fitting residuals, enabling detection of spectral irregularities and unstable bands. Spatial consistency is analysed through local variance statistics, entropy measures, and spatial autocorrelation metrics such as Moran's I and Geary's C. Illumination consistency is evaluated through global and local intensity gradients together with low-frequency spatial components extracted through filtering and Fourier analysis. In addition, covariance and correlation matrices between spectral bands model redundancy and intrinsic spectral dimensionality, while PCA-based descriptors quantify explained variance and eigenvalue decay. Dead and saturated bands are identified through variance and dynamic-range criteria. General outlier descriptors are computed according to the procedures defined in Section 2.

CatBoost was selected because the high-dimensional spectral feature space contains complex non-linear relationships that are effectively captured by gradient-boosting tree ensembles. The resulting model achieved an AUC of approximately 0.92 (Table 2).

In line with the feedback logic described in Section 2, the HSI stage produces a structured set of parameter updates and acquisition flags (e.g., accept, adjust, reacquire, reject), which can be used by the operator or external control modules (Fig. 1). The combined severity score provides effective separation between decision classes, with a clear progression from acceptable to human-intervention cases.

**Table 2**. Performance evaluation of the HSI classification model.

| Metric | Accuracy | Precision | Recall | F1 | AUC |
|---|---|---|---|---|---|
| Value | 0.86 | 0.77 | 0.89 | 0.82 | 0.92 |

# 6. XRF

Each acquisition consists of one or more spectra representing photon counts as a function of X-ray energy, transformed into a structured feature representation combining deterministic quality indicators and higher-order statistical descriptors. Signal-to-noise ratio (SNR) is estimated by comparing peak intensities with local background fluctuations, both at individual peaks and across the spectrum, while total count rate provides a global indicator of statistical quality, with low-count spectra typically associated with noisy peaks, low detection limit for some elements or poor acquisition geometry.

Additional physically grounded indicators are incorporated. Argon peak intensity at 2.96 keV is used as an indirect measure of the detector-to-sample distance, since air gaps increase argon fluorescence and attenuate the useful signal. Peak detectability and prominence are evaluated through local maxima detection and parametric peak fitting methods, spectral baseline stability through polynomial fitting, morphological filtering, and asymmetric least-squares correction, and energy-calibration consistency through the relative stability of detected peak positions across repeated measurements.

These indicators are combined with peak-fitting parameters, peak ratios, baseline-corrected spectral descriptors, and the general statistical and outlier features defined in Section 2. XGBoost achieved the best performance, with an AUC of approximately 0.93 (Table 3).

Following the feedback logic described in Section 2, the XRF stage additionally supports data efficiency and resource-aware operations: spectra identified as unreliable or severely degraded can be excluded from further analysis, while marginal cases may be retained only in processed form (e.g., baseline-corrected or denoised spectra).

**Table 3.** Performance evaluation of the XRF classification model (Catboost).

| Metric | Accuracy | Precision | Recall | F1 | AUC |
|---|---|---|---|---|---|
| Value | 0.94 | 0.63 | 0.65 | 0.64 | 0.93 |

# 7. Raman spectroscopy

Each Raman acquisition consists of one or more spectra representing intensity as a function of Raman shift. Within the general framework of Section 2, these spectra are characterised using a structured feature space combining deterministic spectral indicators and statistical descriptors.

Signal-to-noise ratio (SNR) is estimated by comparing the intensity of characteristic Raman peaks with local noise levels derived from high-frequency spectral components, smoothing residuals, or peak-free spectral regions. Fluorescence background is quantified using polynomial fitting, asymmetric least-squares correction, and morphological filtering. Spectral smoothness and peak sharpness are evaluated using derivative-based measures. Peak detectability and prominence are extracted through local maxima analysis and model-based peak fitting, while baseline stability is assessed through residual-noise statistics after baseline correction. Spectral alignment consistency is additionally monitored through cross-correlation and peak-matching techniques. Detector saturation and dynamic-range utilisation are analysed through intensity-range statistics and threshold-based detection of underexposed or saturated spectra. General outlier descriptors are computed as described in Section 2.
Random Forest, XGBoost, and CatBoost models were trained on labelled spectra. The best-performing models achieve an AUC of approximately 0.83 (see table 4). Although lower than the values obtained for HSI and XRF, these results remain significant given the intrinsic complexity and variability of Raman acquisitions.
Interpretability is introduced through explainable artificial intelligence (XAI) techniques based primarily on SHAP feature attribution analysis. Global feature-importance rankings and local explanations indicate that the classification process is dominated by fluorescence-related descriptors, low-signal and residual-noise indicators, and peak sharpness metrics.
In the feedback stage, corrective actions are prioritised according to severity. For Raman spectroscopy, this distinction is critical because not all degraded spectra are equally actionable: low signal or moderate baseline issues may be corrected through parameter adjustment or reacquisition, whereas severe fluorescence, missing diagnostic peaks, or unclassified peak positions usually require expert review.

**Table 4**. Performance evaluation of the Raman classification model (XGBoost). The results show solid discriminative ability (AUC = 0.83) and satisfactory accuracy, while the moderate Precision, lower Recall, and F1-score suggest limitations in capturing all positive instances and potential class imbalance effects in currently available data.

| Metric | Accuracy | Precision | Recall | F1 | AUC |
|---|---|---|---|---|---|
| Value | 0.80 | 0.73 | 0.61 | 0.66 | 0.83 |

# 8. Automatic calibration loop and final comments

The proposed framework is designed as an adaptive closed-loop architecture in which acquisition, statistical analysis, and control can be continuously integrated. In its current implementation, the framework provides decision support by identifying acquisition degradations and associating them with candidate corrective actions, thereby establishing the basis for future closed-loop robotic calibration. The present work evaluates the proposed methodology as an integrated framework. A quantitative assessment of the contribution of individual components through dedicated ablation studies is left for future investigation.
At the beginning of each acquisition session, reference measurements acquired under nominal conditions are used to estimate empirical distributions describing expected system behaviour. New acquisitions are transformed into structured feature representations and evaluated against these reference distributions using statistical thresholding and machine-learning-based classification. Deviations from expected behaviour, including anomalous clustering, unstable feature distributions, or out-of-bound measurements, are interpreted as indicators of calibration drift, acquisition instability, or reconstruction and spectral degradation.

A central component of the framework is the mapping between feature-space deviations and corrective actions. Specific degradation patterns are associated with adjustments of acquisition parameters such as illumination, exposure time, sensor gain, acquisition geometry, or measurement duration. Updated acquisitions are then reintroduced into the pipeline, enabling iterative refinement of acquisition conditions. Severe or ambiguous cases exceeding acceptable confidence bounds are escalated to human intervention. Fig. 1 illustrates the role of the XAI stage within the proposed calibration loop. By highlighting the contribution of individual quality indicators to each prediction, the framework provides interpretable diagnostic information that links statistical deviations to actionable corrective measures.

The framework is inherently adaptive: as numerous new acquisitions are collected across different sessions and operational conditions, the statistical descriptors, reference distributions, and classification models are progressively updated. This continuous learning process improves robustness and generalisation capability while allowing the system to adapt to evolving acquisition conditions and sensing configurations.

Future work will focus on extending validation using substantially larger datasets acquired under controlled robotic conditions, enabling more robust statistical assessment and refinement of the proposed machine-learning models. It will also include experimental validation of the proposed feedback loop within the robotic acquisition platform, assessing the effectiveness of the suggested corrective actions in improving acquisition quality under real operating conditions. Furthermore, failure-mode classification will be refined beyond the current binary labels, and the diagnostic outputs will be integrated directly into the robotic control layer, completing the transition from an AI-assisted quality assessment framework to a fully adaptive robotic acquisition workflow.

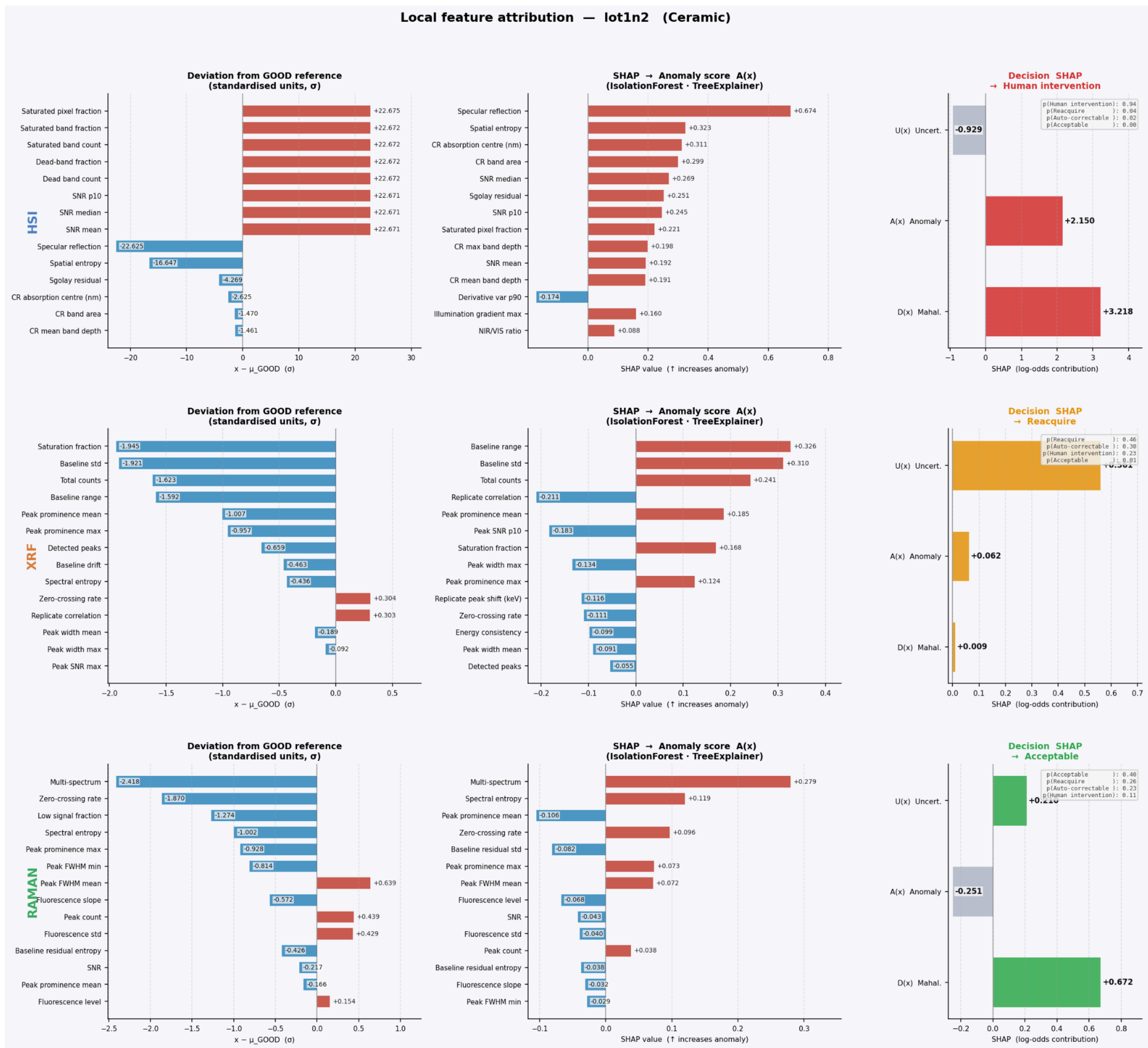

**Fig. 1.** Representative local SHAP explanations illustrating how the proposed calibration framework links classification decisions to physically meaningful quality indicators. Positive and negative feature contributions reveal the causes of acquisition degradation and provide interpretable guidance for selecting corrective actions within the adaptive calibration loop.

# Funding

This research was supported by the European Union's Horizon Europe Research and Innovation Actions through the AUTOMATA project (Grant Agreement No. 101158046), and by UK Research and Innovation (UKRI) under the UK Government's Horizon Europe Funding Guarantee (Grant Nos. 10110150 and 10139794). Views and opinions expressed are, however, those of the author(s) only and do not necessarily reflect those of the European Union, the European Commission, or UKRI. Neither the European Union, the European Commission, nor UKRI can be held responsible for them.